\documentclass[10pt,twocolumn,letterpaper]{article}

\usepackage{iccv}
\usepackage{times}
\usepackage{epsfig}
\usepackage{graphicx}
\usepackage{amsmath}
\usepackage{amssymb}
\usepackage{url}
\usepackage{amsfonts}
\usepackage{subfigure}
\usepackage{multirow}
\usepackage{makecell}
\usepackage{booktabs}
\usepackage{dsfont}

\usepackage[pagebackref=true,breaklinks=true,letterpaper=true,colorlinks,bookmarks=false]{hyperref}

\iccvfinalcopy 

\def\iccvPaperID{675} 

\ificcvfinal\fi
\begin{document}

\title{Multi-modal Factorized Bilinear Pooling with Co-Attention Learning \\
for Visual Question Answering}

\author{Zhou Yu$^\dag$,~~Jun Yu$^\dag$\thanks{Jun Yu is the corresponding author},~~Jianping Fan$^\ddag$,~~Dacheng Tao$^\S$\\
$^\dag$ Key Laboratory of Complex Systems Modeling and Simulation, \\
School of Computer Science and Technology, Hangzhou Dianzi University, P. R. China\\
$^\ddag$ Department of Computer Science, University of North Carolina at Charlotte, USA\\
$^\S$ UBTECH Sydney AI Centre, School of IT, FEIT, The University of Sydney, Australia\\
{\tt\small yuz@hdu.edu.cn, yujun@hdu.edu.cn, jfan@uncc.edu, dacheng.tao@sydney.edu.au}
}

\maketitle

\begin{abstract}
Visual question answering (VQA) is challenging because it requires a simultaneous understanding of both the visual content of images and the textual content of questions. The approaches used to represent the images and questions in a fine-grained manner and questions and to fuse these multi-modal features play key roles in performance. Bilinear pooling based models have been shown to outperform traditional linear models for VQA, but their high-dimensional representations and high computational complexity may seriously limit their applicability in practice. For multi-modal feature fusion, here we develop a Multi-modal Factorized Bilinear (MFB) pooling approach to efficiently and effectively combine multi-modal features, which results in superior performance for VQA compared with other bilinear pooling approaches. For fine-grained image and question representation, we develop a `co-attention' mechanism using an end-to-end deep network architecture to jointly learn both the image and question attentions. Combining the proposed MFB approach with co-attention learning in a new network architecture provides a unified model for VQA. Our experimental results demonstrate that the single MFB with co-attention model achieves new state-of-the-art performance on the real-world VQA dataset. Code available at \url{https://github.com/yuzcccc/mfb}.
\end{abstract}

\section{Introduction}
Thanks to recent advances in computer vision and natural language processing, computers are expected to be able to automatically understand the semantics of images and natural languages in the near future. Such advances have also stimulated new research topics like image-text retrieval \cite{yang2008harmonizing,yu2014discriminative}, image captioning \cite{donahue2015long,xu2015show}, and visual question answering \cite{antol2015vqa,malinowski2014multi}.

Compared with image-text retrieval and image captioning (which just require the underlying algorithms to {search} or {generate} a free-form text description for a given image), \emph{visual question answering} (VQA) is a more challenging task that requires fine-grained understanding of the semantics of both the images and the questions as well as supports complex reasoning to predict the best-matching answer correctly. In some aspects, the VQA task can be treated as a generalization of image captioning and image-text retrieval. Thus building effective VQA algorithms, which can achieve close performance like human beings, is an important step towards enabling artificial intelligence in general.

Existing VQA approaches usually have three stages: (1) representing the images as visual features and questions as textual features; (2) combining these multi-modal features to obtain fused image-question features; (3) using the integrated image-question features to learn a multi-class classifier and to predict the best-matching answer. Deep neural networks (DNNs) are effective and flexible, many existing approaches model the three stages in one DNN model and train the model in an end-to-end fashion through back-propagation. In the three stages, {feature representation} and {multi-modal feature fusion} particular affect VQA performance.

With respect to multi-modal feature fusion, most existing approaches simply use linear models for multi-modal feature fusion (e.g., concatenation or element-wise addition) to integrate the image's visual feature with the question's textual feature \cite{zhou2015simple,lu2016hierarchical}. Since multi-modal feature distributions may vary dramatically, the integrated image-question representations obtained by such linear models may not be sufficiently expressive to fully capture complex associations between the visual features from images and the textual features from questions. In contrast to linear pooling, bilinear pooling \cite{tenenbaum1997separating} has recently been used to integrate different CNN features for fine-grained image recognition \cite{lin2015bilinear}. However, the high dimensionality of the output features and the huge number of model parameters may seriously limit the applicability of bilinear pooling. Fukui \emph{et al.} proposed the Multi-modal Compact Bilinear (MCB) pooling model to effectively and simultaneously reduce the number of parameters and computation time using the Tensor Sketch algorithm \cite{fukui2016multimodal}. Using the MCB model, the group proposed a network architecture for the VQA task and won the VQA challenge 2016. Nevertheless, the MCB model lies on a high-dimensional output feature to guarantee robust performance, which may limit its applicability due to huge memory usage. To overcome this problem, Kim \emph{et al.} proposed the Multi-modal Low-rank Bilinear (MLB) pooling model based on the Hadamard product of two feature vectors \cite{kim2016hadamard}. Since MLB generate output features with lower dimensions and models with fewer parameters, it is highly competitive with MCB. However, MLB has a slow convergence rate and is sensitive to the learned hyper-parameters. To address these issues, here we develop the Multi-modal Factorized Bilinear pooling (MFB) method, which enjoys the dual benefits of compact output features of MLB and robust expressive capacity of MCB.

With respect to feature representation, directly using global features for image representation may introduce noisy information that is irrelevant to the given question. Therefore, it is intuitive to introduce \emph{visual attention} mechanism \cite{xu2015show} into the VQA task to adaptively learn the most relevant image regions for a given question. Modeling visual attention may significantly improve performance \cite{fukui2016multimodal}. However, most existing approaches only model image attention without considering question attention, even though question attention is also very important since the questions interpreted in natural languages may also contain colloquialisms that can be regarded as noise. Therefore, based on our MFB approach, we design a deep network architecture for the VQA task using a \emph{co-attention learning} module to jointly learn both image and question attentions.

To summarize, the main contributions of this study are as follows: First, we develop a simple but effective Multi-modal Factorized Bilinear pooling (MFB) approach to fuse the visual features from images with the textual features from questions. MFB significantly outperforms existing multi-modal bilinear pooling approaches such as MCB \cite{fukui2016multimodal} and MLB \cite{kim2016hadamard}. Second, based on the MFB module, a \emph{co-attention} learning architecture is designed to jointly learn both image and question attention. Our MFB approach with co-attention model achieves the state-of-the-art performance on the VQA dataset. We also conduct detailed and extensive experiments to show why our MFB approach is effective. Our experimental results demonstrate that normalization techniques are extremely important in bilinear models.

\section{Related Work}\label{sec:related_work}
In this section, we briefly review the most relevant research on VQA, especially those studies that use multi-modal bilinear models.

\subsection{Visual Question Answering (VQA)}
Malinowski \emph{et al.} \cite{malinowski2014multi} made an early attempt at solving the VQA task. Since then, solving the VQA task has received increasing attention from the computer vision and natural language processing communities. VQA approaches can be classified into the following methodological categories: the coarse joint-embedding models \cite{zhou2015simple,antol2015vqa,kim2016multimodal,saito2016dualnet}, the fine-grained joint-embedding models with attention \cite{andreas2016learning,lu2016hierarchical,fukui2016multimodal,ilievski2016focused,nam2016dual,zhao2017video,zhu2015uncovering} and the external knowledge based models \cite{wang2015explicit,wang2016fvqa,wu2016ask}.

The coarse joint-embedding models are the most straightforward VQA solutions. Image and question are first represented as global features and then integrated to predict the answer. Zhou \emph{et al.} proposed a baseline approach for the VQA task by using the concatenation of the image CNN features and the question BoW (bag-of-words) features, with a linear classifier learned to predict the answer \cite{zhou2015simple}. Some approaches introduce more complex deep models, e.g., LSTM networks \cite{antol2015vqa} or residual networks \cite{kim2016multimodal}, to tackle the VQA task in an end-to-end fashion.

One limitation of coarse joint-embedding models is that their global features may contain noisy information, making it hard to correctly answer fine-grained problems (e.g., ``what color are the cat's eyes?'') . Therefore, recent VQA approaches introduce the \emph{visual attention} mechanism \cite{xu2015show} into the VQA task by adaptively learning the local fine-grained image features for a given question. Chen \emph{et al.} proposed a ``question-guided attention map'' that projects the question embeddings to the visual space and formulates a configurable convolutional kernel to search the image attention region \cite{chen2015abc}. Yang \emph{et al.} proposed a stacked attention network to learn the attention iteratively \cite{yang2016stacked}. Some approaches introduce off-the-shelf object detectors \cite{ilievski2016focused} or object proposals \cite{shih2016look} as the attention region candidates and then use the question to identify related ones. Fukui \emph{et al.} proposed multi-modal compact bilinear pooling to integrate image features from spatial grids with textual features from the questions to predict the attention \cite{fukui2016multimodal}. In addition, some approaches apply attention learning to both the images and questions. Lu \emph{et al.} proposed a co-attention learning framework to alternately learn the image attention and the question attention \cite{lu2016hierarchical}. Nam \emph{et al.} proposed a multi-stage co-attention learning framework to refine the attentions based on memory of previous attentions \cite{nam2016dual}.

Despite joint embedding models for VQA delivering impressive performance, they are not good enough for answering problems that require complex reasoning or common sense knowledge. Therefore, introducing external knowledge is beneficial for VQA. However, existing approaches have either only been applied to specific datasets \cite{wang2015explicit,wang2016fvqa}, or have been ineffective on benchmark datasets \cite{wu2016ask}. There is room for further exploration and development.

\subsection{Multi-modal Bilinear Models for VQA}
Multi-modal feature fusion plays an important and fundamental role in VQA. After the image and question features are obtained, concatenation or element-wise summations are most frequently used for multi-modal feature fusion. Since the distributions of two feature sets in different modalities (i.e.,the visual features from images and the textual features from questions) may vary significantly, the representation capacity of the fused features may be insufficient, limiting the final prediction performance.

Fukui \emph{et al.} first introduced the bilinear model to solve the problem of multi-modal feature fusion in VQA. In contrast to the aforementioned approaches, they proposed the Multi-modal Compact Bilinear pooling (MCB), which uses the outer product of two feature vectors to produce a very high-dimensional feature for quadratic expansion \cite{fukui2016multimodal}. To reduce the computational cost, they used a sampling-based approximation approach that exploits the property that the projection of two vectors can be represented as their convolution. The MCB model outperformed the simple fusion approaches and demonstrated superior performance on the VQA dataset \cite{antol2015vqa}. Nevertheless, MCB usually needs high-dimensional features (e.g., 16,000-D) to guarantee robust performance, which may seriously limit its applicability due to limitations in GPU memory.

To overcome this problem, Kim \emph{et al.} proposed the Multi-modal Low-rank Bilinear Pooling (MLB) approach based on the Hadamard product of two feature vectors (i.e., the image feature $x\in\mathbb{R}^m$ and the question feature $y\in\mathbb{R}^n$) in the common space with two low-rank projection matrices: \cite{kim2016hadamard}:
\begin{equation}\label{eq:mlb}
z = \mathrm{MLB}(x,y) = (U^Tx)\circ (V^Ty)
\end{equation}
where $U\in\mathbb{R}^{m\times o}$ and $V\in\mathbb{R}^{n\times o}$ are the projection matrices, $o$ is the dimensionality of the output feature, and $\circ$ denotes the Hadamard product or the element-wise multiplication of two vectors. To further increase model capacity, nonlinear activation like $tanh$ is added after $z$. Since the MLB approach can generate feature vectors with low dimensions and deep models with fewer parameters, it has achieved comparable performance to MCB. In \cite{kim2016hadamard}, the experimental results indicated that MLB may lead to a slow convergence rate (the MLB with attention model takes 250k iterations, which is about 140 epochs, to converge \cite{kim2016hadamard}).

\section{Multi-modal Factorized Bilinear Pooling}\label{sec:mfb}

Given two feature vectors in different modalities, e.g., the visual features $x\in\mathbb{R}^{m}$ for an image and the textual features $y\in\mathbb{R}^{n}$ for a question, the simplest multi-modal bilinear model is defined as follows:
\begin{equation}\label{eq:multimodal_bilinear_base}
z_i = x^TW_iy
\end{equation}
where $W_i\in\mathbb{R}^{m\times n}$ is a projection matrix, $z_i\in\mathbb{R}$ is the output of the bilinear model. The bias term is omitted here since it is implicit in $W$. To obtain a $o$-dimensional output $z$, we need to learn $W=[W_i,...,W_o]\in\mathbb{R}^{m \times n \times o}$. Although bilinear pooling can effectively capture the pairwise interactions between the feature dimensions, it also introduces huge number of parameters that may lead to high computational cost and a risk of over-fitting.

Inspired by the matrix factorization tricks for uni-modal data \cite{li2016factorized,rendle2010factorization}, the projection matrix $W_i$ in Eq.(\ref{eq:multimodal_bilinear_base}) can be factorized as two low-rank matrices:
\begin{equation}\label{eq:mfb}
\begin{array}{rcl}
z_i &=& x^TU_iV_i^Ty = \sum\limits_{d=1}^kx^Tu_dv_d^Ty\\
&=& \mathds{1}^T(U_i^Tx\circ V_i^Ty)
\end{array}
\end{equation}
where $k$ is the factor or the latent dimensionality of the factorized matrices $U_i=[u_1,...,u_k]\in\mathbb{R}^{m \times k}$ and $V_i=[v_1,...,v_k]\in\mathbb{R}^{n \times k}$, $\circ$ is the Hadmard product or the element-wise multiplication of two vectors, $\mathds{1}\in\mathbb{R}^k$ is an all-one vector.

To obtain the output feature $z\in\mathbb{R}^o$ by Eq.(\ref{eq:mfb}), the weights to be learned are two three-order tensors $U=[U_1,...,U_o]\in\mathbb{R}^{m\times k \times o}$ and $V=[V_1,...,V_d]\in\mathbb{R}^{n\times k \times o}$ accordingly. Without loss of generality, we can reformulate $U$ and $V$ as 2-D matrices $\tilde{U}\in\mathbb{R}^{m \times ko}$ and $\tilde{V}\in\mathbb{R}^{n \times ko}$ respectively with simple reshape operations. Accordingly, Eq.(\ref{eq:mfb}) can be rewritten as follows:
\begin{equation}\label{eq:mfb_matrix}
z = \mathrm{SumPooling}(\tilde{U}^Tx\circ \tilde{V}^Ty, k)
\end{equation}
where the function $\mathrm{SumPooling}(x,k)$ means using a one-dimensional non-overlapped window with the size $k$ to perform sum pooling over $x$. We name this model Multi-modal Factorized Bilinear pooling (MFB).

\begin{figure}
\begin{center}
\subfigure[Multi-modal Factorized Bilinear Pooling] {\includegraphics[width=0.62\linewidth]{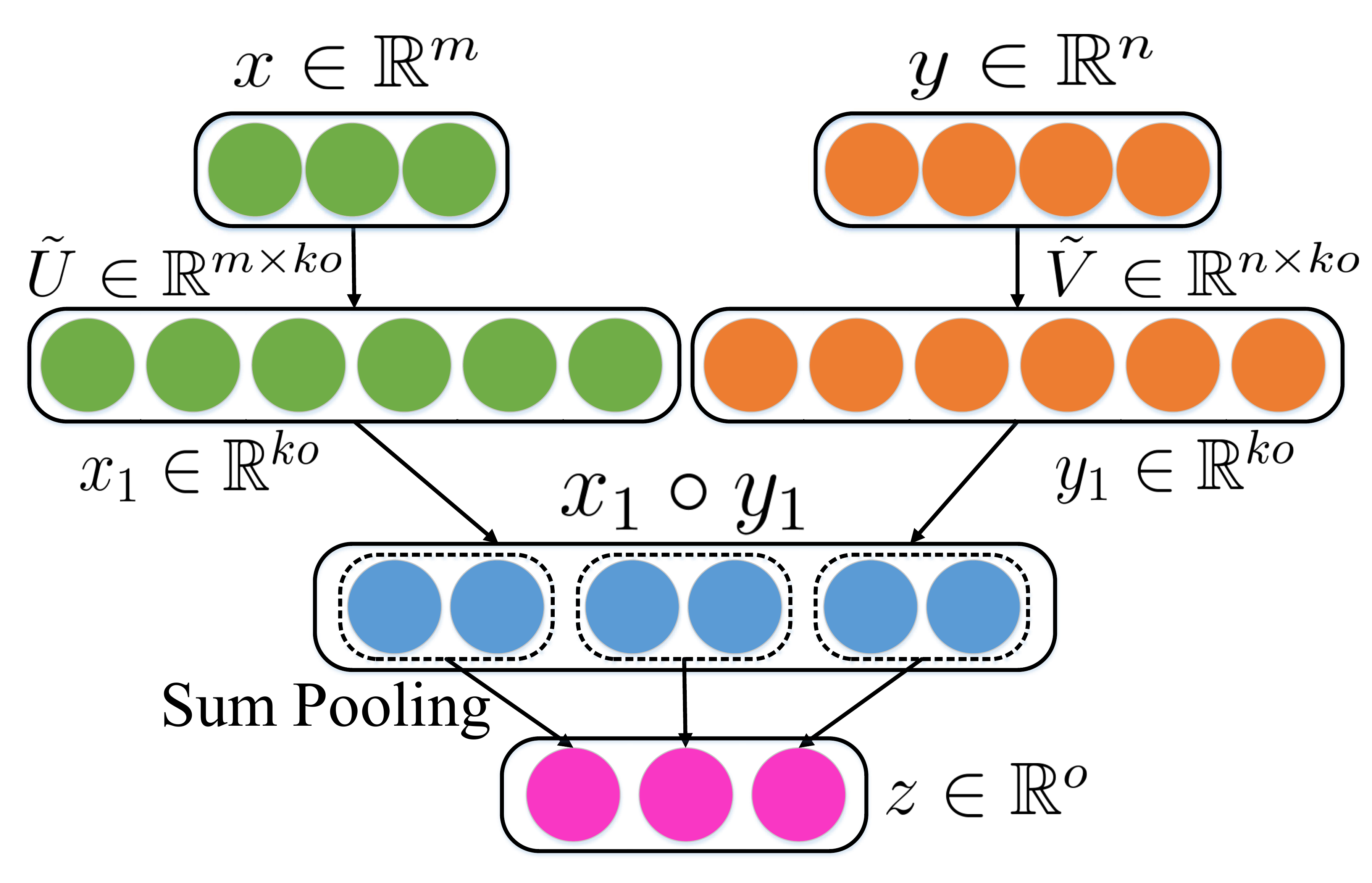}\label{fig:mfb}}
\subfigure[MFB module] {\includegraphics[width=0.37\linewidth]{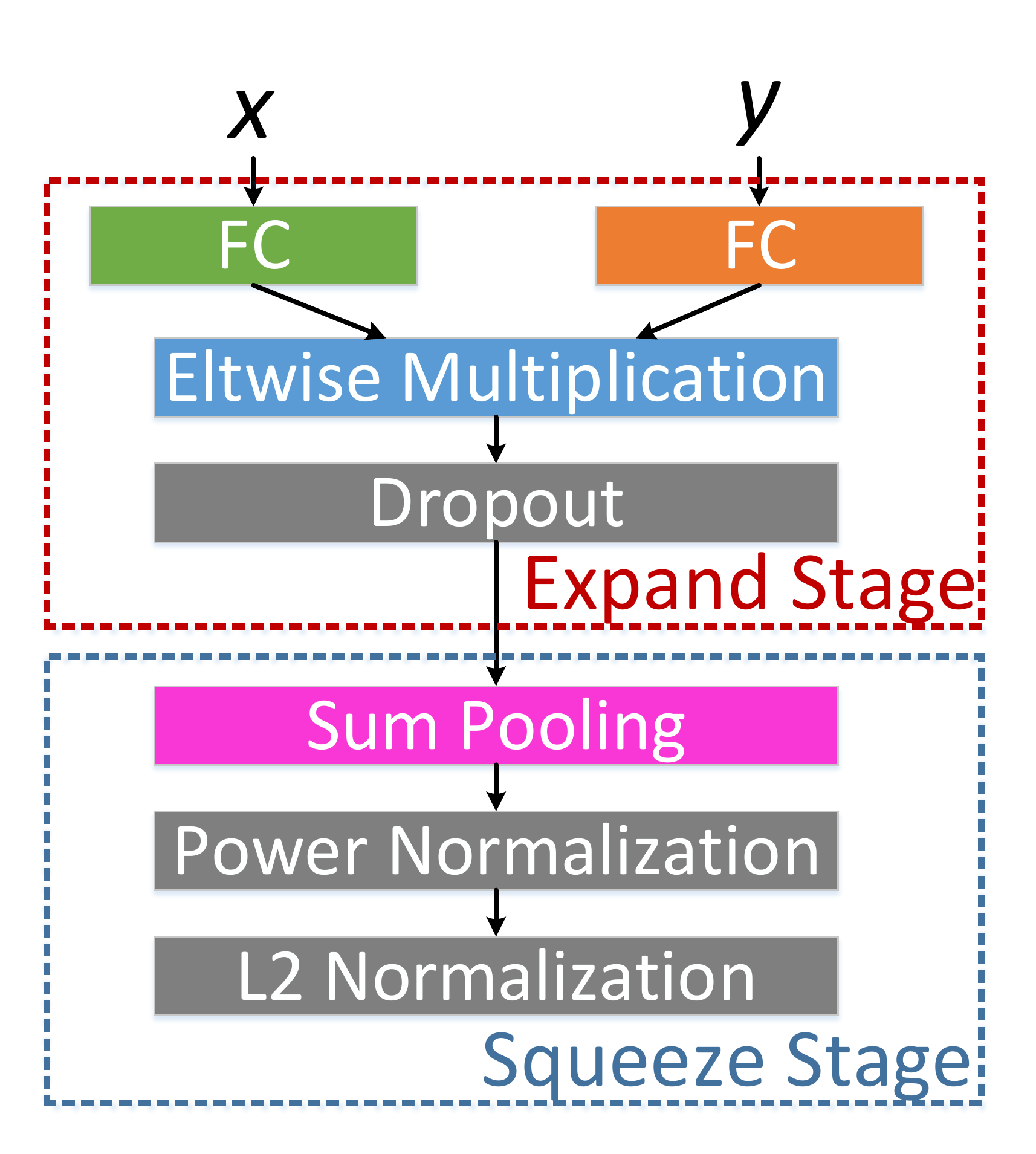}\label{fig:mfb_module}}
\caption{The flowchart of Multi-modal Factorized Bilinear Pooling and completed design of the MFB module.}
\label{fig:mfb_framework}
\end{center}
\vspace{-15pt}
\end{figure}

The detailed procedures of MFB are illustrated in Fig. \ref{fig:mfb}. The approach can be easily implemented by combining some commonly-used layers such as fully-connected, element-wise multiplication and pooling layers. Furthermore, to prevent over-fitting, a dropout layer is added after the element-wise multiplication layer. Since element-wise multiplication is introduced, the magnitude of the output neurons may vary dramatically, and the model might converge to an unsatisfactory local minimum. Therefore, similar to \cite{fukui2016multimodal}, the power normalization ($z \leftarrow \mathrm{sign}(z)|z|^{0.5}$) and $\ell_2$ normalization ($z \leftarrow z/\|z\|$) layers are appended after MFB output. The flowchart of the entire MFB module is illustrated in Fig. \ref{fig:mfb_module}.

\textbf{Relationship to MLB. }
Eq.(\ref{eq:mfb_matrix}) shows that the MLB in Eq.(\ref{eq:mlb}) is a special case of the proposed MFB with $k=1$, which corresponds to the rank-1 factorization. Figuratively speaking, MFB can be decomposed into two stages (see in Fig. \ref{fig:mfb_module}): first, the features from different modalities are \emph{expanded} to a high-dimensional space and then integrated with element-wise multiplication. After that, sum pooling followed by the normalization layers are performed to \emph{squeeze} the high-dimensional feature into the compact output feature, while MLB directly projects the features to the low-dimensional output space and performs element-wise multiplication. Therefore, with the same dimensionality for the output features,
the representation capacity of MFB is more powerful than MLB.

\section{Network Architectures for VQA}\label{sec:nn_arch}
The goal of the VQA task is to answer a question about an image. The inputs to the model contain an image and a corresponding question about the image. Our model extracts both the image and the question representations, integrates the multi-modal features using the MFB module in Figure \ref{fig:mfb_module}, treats each individual answer as one class and performs multi-class classification to predict the correct answer. In this section, two network architectures are introduced. The first is the MFB baseline with one MFB module, which is used to perform ablation analysis with different hyper-parameters for comparison with other baseline approaches. The second network introduces co-attention learning which jointly learns the image and question attentions, to better capture fine-grained correlations between the image and the question, which may lead to a model with better representation capability.
\begin{figure}
\begin{center}
\includegraphics[width=0.49\textwidth]{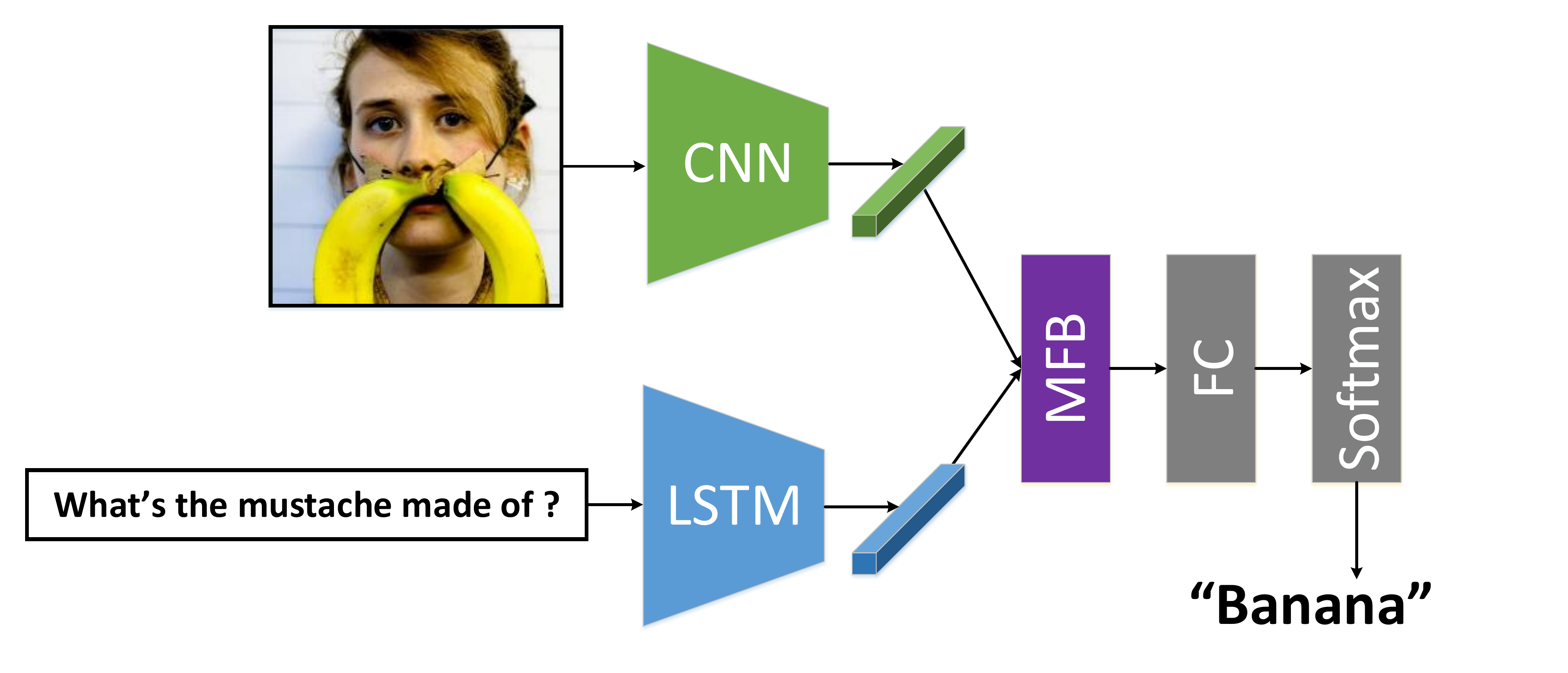}
\caption{MFB baseline network architecture for VQA.}
\label{fig:mfb_baseline}
\end{center}
\vspace{-15pt}
\end{figure}

\begin{figure*}
\begin{center}
\includegraphics[width=1\textwidth]{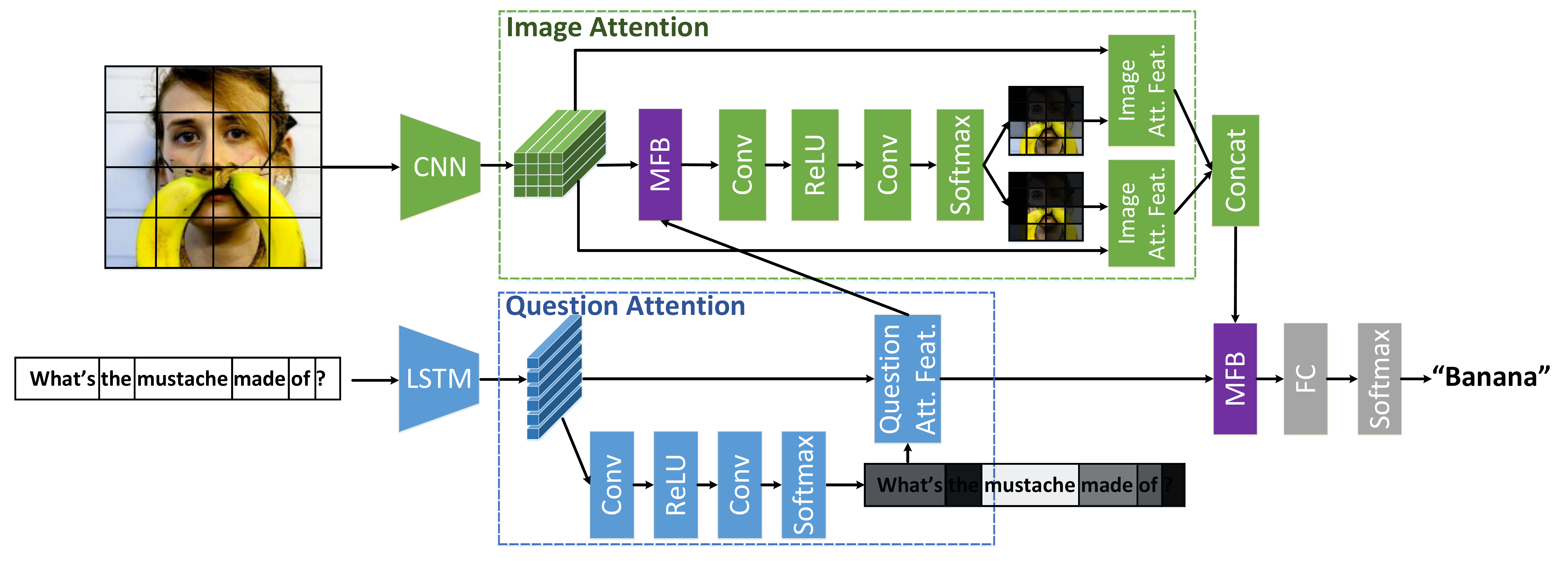}
\caption{MFB with Co-Attention network architecture for VQA. Different from the network of MFB baseline, the images and questions are firstly represented as the fine-grained features respectively. Then, \emph{Question Attention} and \emph{Image Attention} modules are jointly modeled in the framework to provide more accurate answer predictions.}
\label{fig:mfb_coatt}
\end{center}
\vspace{-15pt}
\end{figure*}
\subsection{MFB Baseline}
Similar to \cite{fukui2016multimodal}, we extract the image features using 152-layer ResNet model \cite{he2015deep} pre-trained on the ImageNet dataset. Images are resized to 448 $\times$ 448, and 2048-D \emph{pool5} features (with $\ell_2$ normalization) are used for image representation. Questions are first tokenized into words, and then further transformed to one-hot feature vectors with max length $T$. Then, the one-hot vectors are passed through an embedding layer and fed into a two-layer LSTM networks with 1024 hidden units \cite{hochreiter1997long}. Each LSTM layer outputs a 1024-D feature for each word. Similar to \cite{fukui2016multimodal}, we extract the output feature of the last word from each LSTM network, and concatenate the obtained features of two LSTM networks to form a 2048-D feature vector for question representation. For predicting the answers, we simply use the top-$N$ most frequent answers as $N$ classes since they follow the long-tail distribution.

The extracted image and question features are fed to the MFB module to generate the fused feature $z$. Finally, $z$ is fed to a $N$-way classifier with the KL-divergence loss. Therefore, all the weights except the ones for the ResNet (due to the limitation of GPU memory) are optimized jointly in an end-to-end manner. The whole network architecture is illustrated in Figure \ref{fig:mfb_baseline}.

\subsection{MFB with Co-Attention}

For a given image, different questions could result in entirely different answers. Therefore, an \emph{image attention} model, which can predict the relevance of each spatial grid to the question, is beneficial for predicting the accurate answer. In \cite{fukui2016multimodal}, 14$\times$14 (196) image spatial grids (\emph{res5c} feature maps in ResNet) are used to represent the input image. After that, the question feature is merged with each of the 196 image features using MCB, followed by some feature transformations (e.g., 1 $\times$ 1 convolution and ReLU activation) and softmax normalization to predict the attention weight for each grid location. Based on the attention map, the attentional image features are obtained by the weighted sum of the spatial grid vectors. Multiple attention maps are generated to enhance the learned attention map, and these attention maps are concatenated to output the attentional image features. Finally, the attentional image features are merged with the question features using MCB to determine the final answer prediction.

From the results reported in \cite{fukui2016multimodal}, one can see that incorporating an attention mechanism allows the model to effectively learn which region is important for the question, clearly contributing to better performance than the model without attention. However, the attention model in \cite{fukui2016multimodal} only focuses on learning image attention while completely ignoring question attention. Since the questions are interpreted as natural language, the contribution of each word is significantly different. Therefore, here we develop a co-attention learning approach (see Figure \ref{fig:mfb_coatt}) to jointly learn both the question and image attentions.

The difference between the network architecture of our co-attention model and the attention model in \cite{fukui2016multimodal} is that we additionally place a \emph{question attention} module after the LSTM networks to learn the attention weights of every word in the question. Different to other co-attention models for VQA \cite{lu2016hierarchical,nam2016dual}, in our model, the image and question modules are loosely coupled such that we do not exploit the image features when learning the question attention module. This is because we assume that the network can directly infer the question attention (i.e., the key words of the question) without seeing the image, as humans do. We name this network MFB with Co-Attention (MFB+CoAtt).

\begin{figure*}
\begin{center}
\subfigure[Standard] {\includegraphics[width=0.24\linewidth]{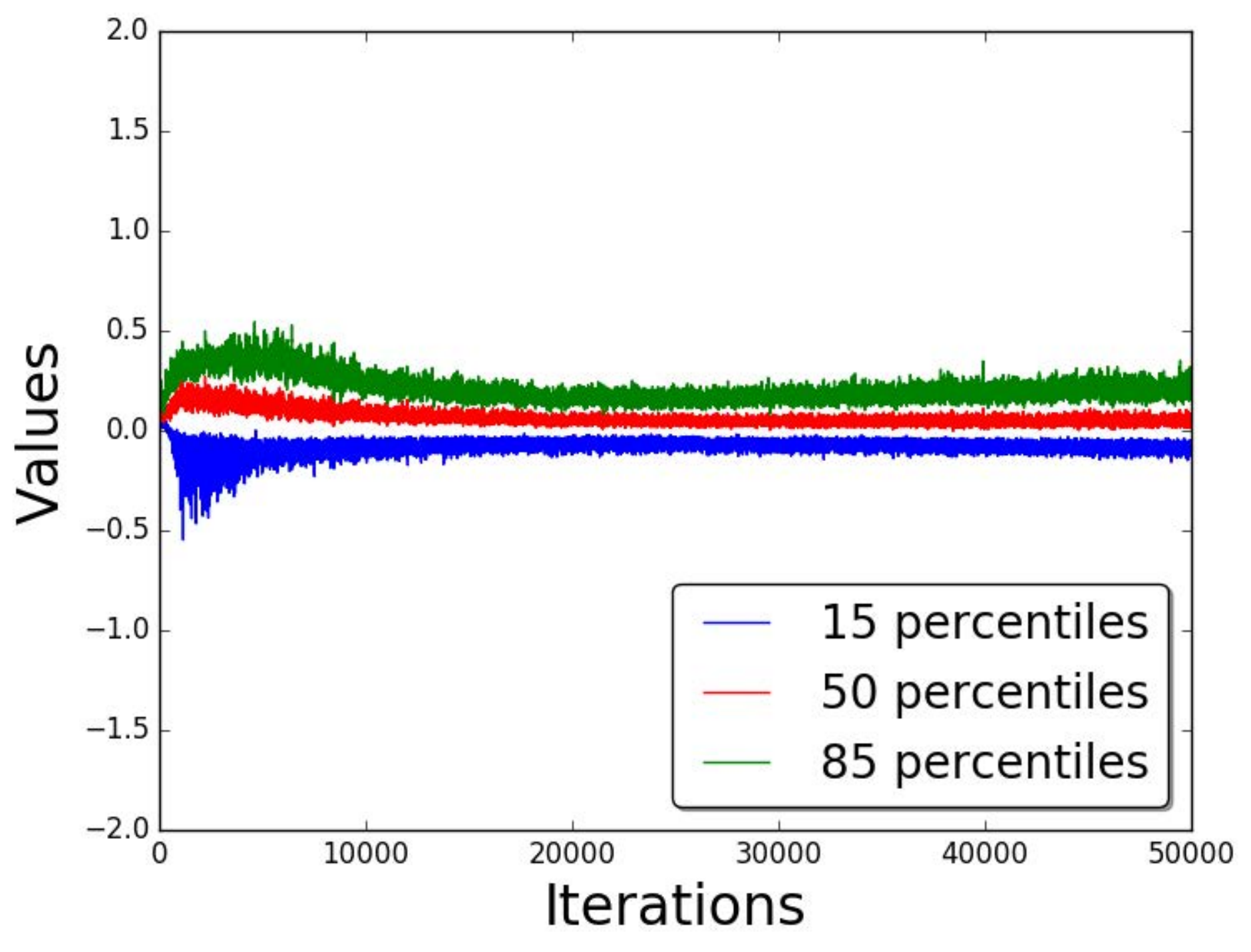}\label{fig:percentile_std}}
\subfigure[w/o power norm.] {\includegraphics[width=0.24\linewidth]{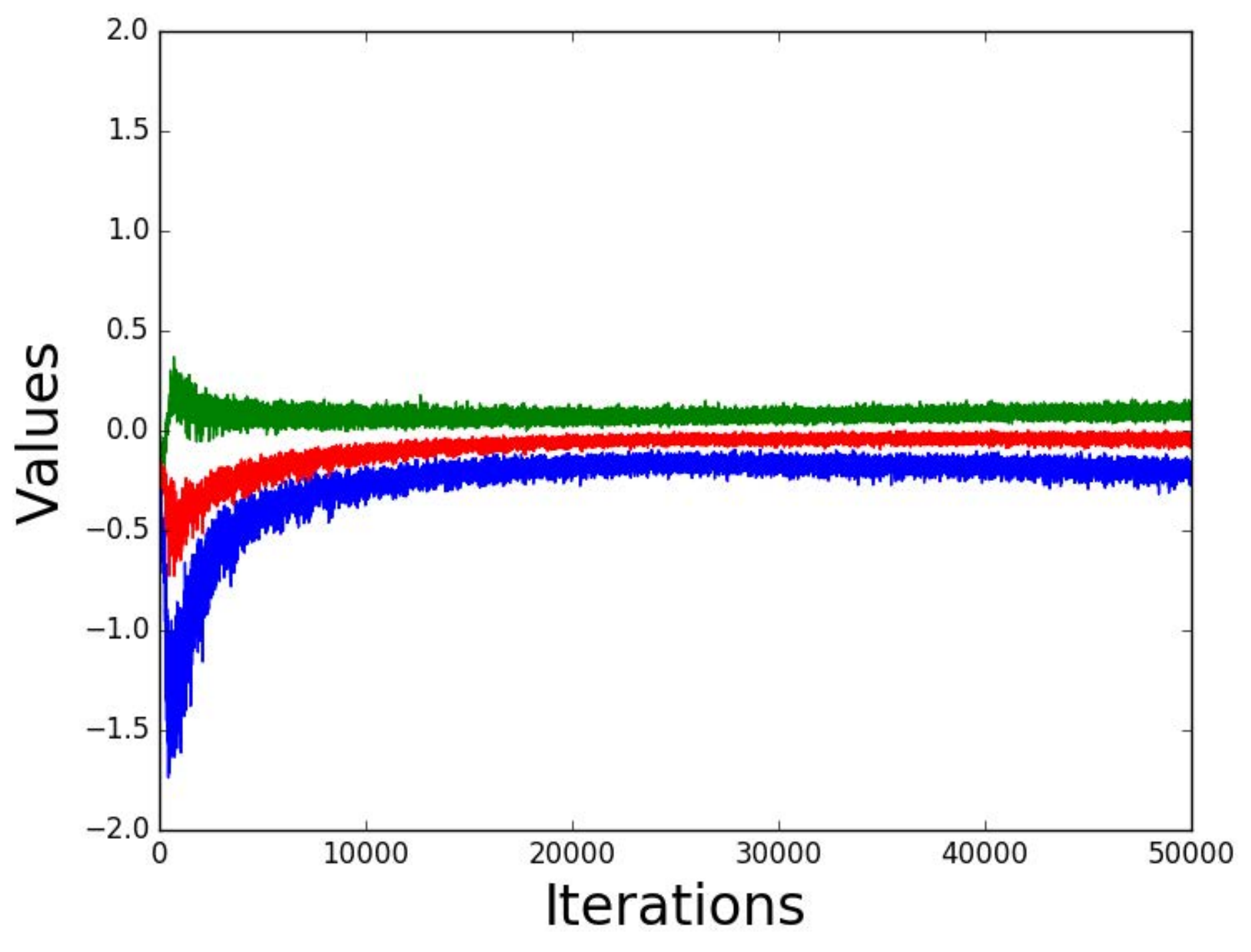}\label{fig:percentile_no_powernorm}}
\subfigure[w/o $\ell_2$ norm.] {\includegraphics[width=0.24\linewidth]{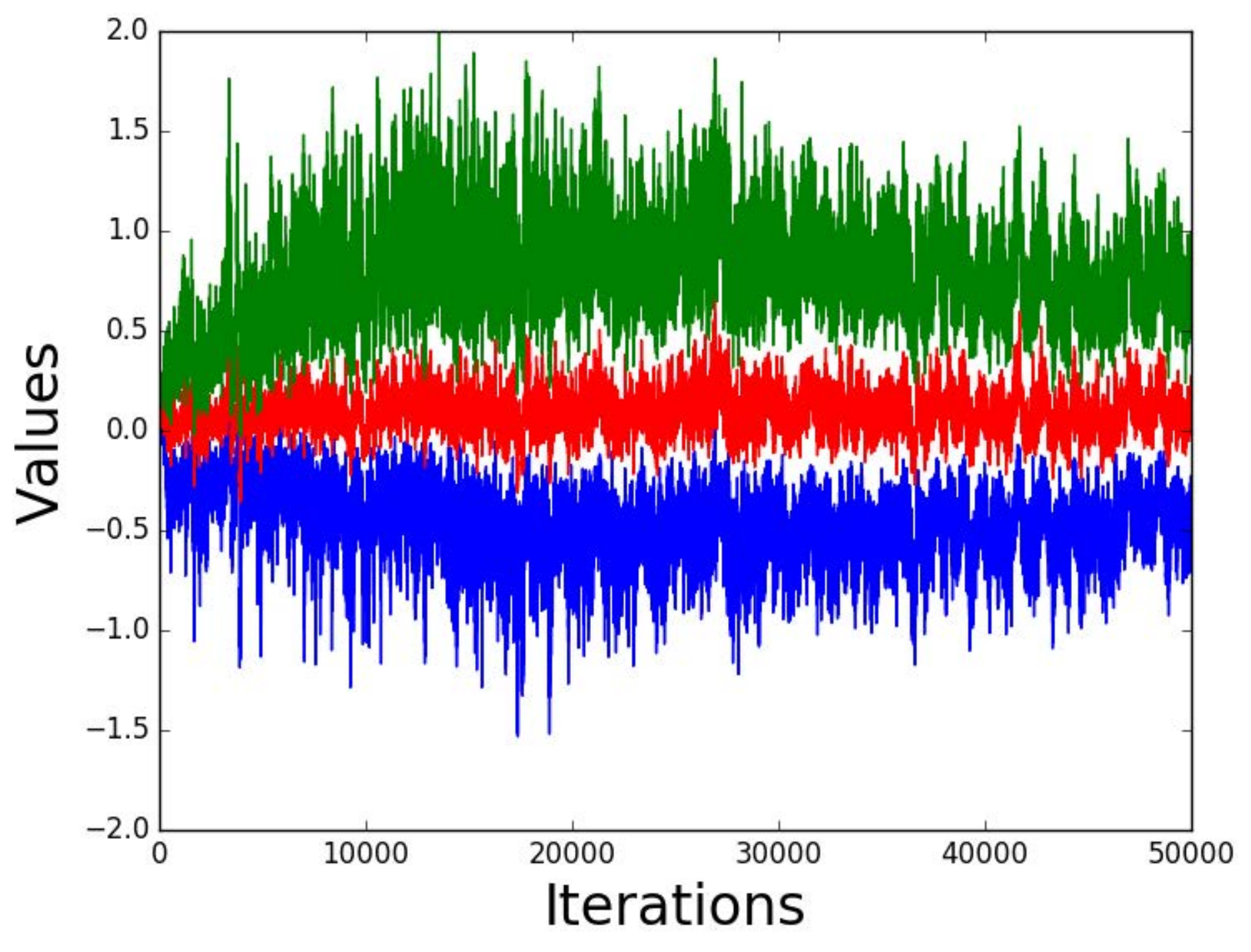}\label{fig:percentile_no_l2norm}}
\subfigure[w/o power and $\ell_2$ norms.] {\includegraphics[width=0.24\linewidth]{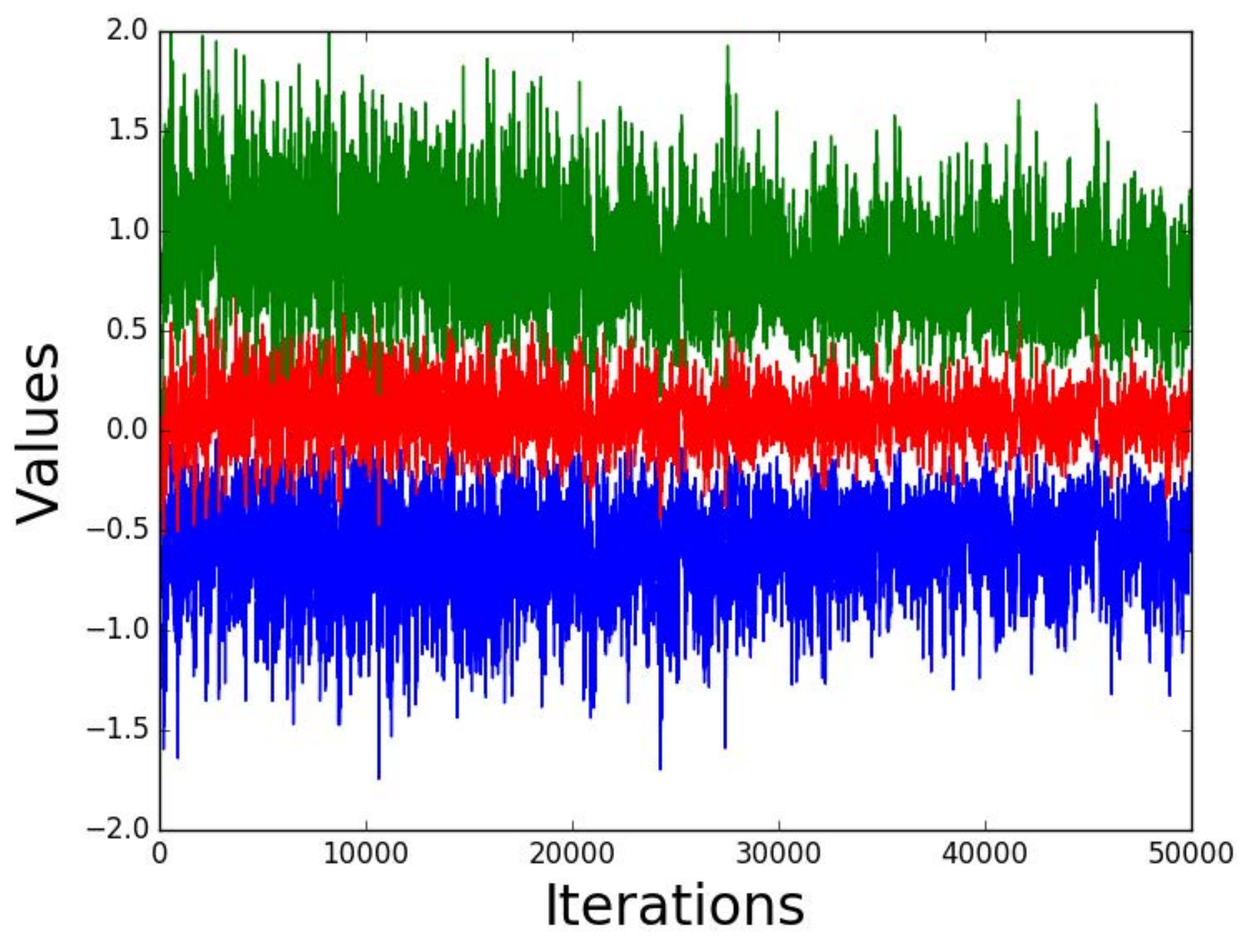}\label{fig:percentile_no_bothnorms}}
\caption{The evolution of the output distribution of one typical neuron with different normalization settings, shown as \{15,50,85\}th percentiles. Both normalization techniques, especially the $\ell_2$ normalization make the neuron values restricted within a narrow range, thus leading to a more stable model. Best viewed in color.}
\label{fig:percentile}
\end{center}
\vspace{-10pt}
\end{figure*}

\section{Experiments}\label{sec:experiments}
In this section, we conduct several experiments to evaluate the performance of our MFB models on the VQA task using the VQA dataset \cite{antol2015vqa} to verify our approach. We first perform ablation analysis on the MFB baseline model to verify the efficiency of the proposed approach over existing state-of-the-art methods such as MCB \cite{fukui2016multimodal} and MLB \cite{kim2016hadamard}. We then provide detailed analyses of the reasons why our MFB model outperforms its counterparts. Finally, we choose the optimal hyper-parameters for the MFB module and train the model with co-attention (MFB+CoAtt) for fair comparison with other state-of-the-art approaches on the VQA dataset \cite{antol2015vqa}.

\subsection{Datasets}
The VQA dataset \cite{antol2015vqa} consists of approximately 200,000 images from the MS-COCO dataset \cite{lin2014microsoft}, with 3 questions per image and 10 answers per question. The data set is split into three: \emph{train} (80k images and 248k questions), \emph{val} (40k images and 122k questions), and \emph{test} (80k images and 244k questions). Additionally, there is a 25$\%$ test split subset named \emph{test-dev}. Two tasks are provided to evaluate performance: Open-Ended (OE) and Multiple-Choices (MC). We use the tools provided by Antol \emph{et al.} \cite{antol2015vqa} to evaluate the performance on the two tasks.

\subsection{Experimental Setup}

For the VQA dataset, we use the Adam solver with $\beta_1=0.9$, $\beta_2=0.99$. The base learning rate is set to 0.0007 and decays every 40,000 iterations using an exponential rate of 0.5. We terminate training at 100,000 iterations (200,000 iterations if the training set is augmented with the large-scale Visual Genome dataset \cite{krishna2016visual}). Dropouts are used after each LSTM layer (dropout ratio $p=0.3$) and MFB module ($p=0.1$) like \cite{fukui2016multimodal}. The number of answers $N=3000$. For all experiments (except for the ones shown in Table \ref{table:sota}, which use the train and val splits together as the training set like the comparative approaches), we train on the {train} split, validate on the {val} split, and report the results on the test split\footnote{the submission attempts for the test-standard split are strictly limited. Therefore, we evaluate most of our settings on the test-dev split and only report the best results on the test-standard split.}. The batch size is set to 200 for the models without the attention mechanism, and set to 64 for the models with attention (due to GPU memory limitation).
All experiments are implemented with the \emph{Caffe} toolbox \cite{jia2014caffe} and performed on a workstation with GTX 1080 GPUs.

\subsection{Ablation Analysis}

\begin{table}
\centering
\caption{Overall accuracies and model sizes of approaches and on the test-dev set of the Open-Ended task. The reported accuracy is the overall accuracy of all question types. The model size includes the parameters for the LSTM networks.
}
\label{table:base}
\begin{tabular}{lccc}
Model & Acc. & Model Size \\
\hline
MCB\cite{fukui2016multimodal} ($d=16000$) & 59.8 & 63M \\
MLB\cite{kim2016hadamard} ($d=1000$) & 59.7 & 25M \\
\hline
MFB($k=1, o=5000$) & 60.4 & 51M \\
MFB($k=5, o=1000$) & \textbf{60.9} & 46M \\
MFB($k=10, o=500$) & 60.6 & 38M \\
\hline
MFB($k=5, o=200$) & 59.8 & 22M \\
MFB($k=5, o=500$) & 60.4 & 28M \\
MFB($k=5, o=2000$) & 60.7 & 62M \\
MFB($k=5, o=4000$) & 60.4 & 107M \\
\hline
MFB($k=5, o=1000$) & - & - \\
~-w/o power norm. & 60.4& - \\
~-w/o $\ell_2$ norm. & 57.7 & - \\
~-w/o power and $\ell_2$ norms. & 57.3 & -\\
\hline
\end{tabular}
\vspace{-10pt}
\end{table}



In Table \ref{table:base}, we compare MFB's performance with other state-of-the-art bilinear pooling models, namely MCB \cite{fukui2016multimodal} and MLB (for fair comparison, we replace the tanh function in MLB with the proposed power+$\ell_2$ normalizations ) \cite{kim2016hadamard}, under the same experimental settings. None of these methods introduce the attention mechanism. Furthermore, we explore different hyper-parameters and normalizations introduced in MFB to explore why MFB outperform the compared bilinear models.

From Table \ref{table:base}, we can see that:

\begin{table*}
\small
\centering
\caption{Open-Ended (OE) and Multiple-Choice (MC) results on VQA dataset compared with the state-of-the-art approaches in terms of accuracy in $\%$. {Att.} indicates whether the approach introduce the attention mechanism, {W.E.} indicates whether the approach uses external word embedding models. VG indicates the model is trained with the Visual Genome dataset additionally. All the reported results are obtained with \emph{a single model}. For the test-dev set, the best results in each split are bolded. For the test-standard set, the best results overall all the splits are bolded.}
\label{table:sota}
\begin{tabular}{l|cc|ccccc|ccccc}

{Model}& {Att.} & {W.E.} & \multicolumn{5}{c|}{{Test-dev}} & \multicolumn{5}{c}{{Test-standard}} \\
\hline
 &&&\multicolumn{4}{c}{{OE}} & {MC} & \multicolumn{4}{c}{{OE}} & {MC}\\
\cline{4-7}
\cline{9-12}
 &&& All & Y/N & Num & Other & All & All & Y/N & Num & Other & All\\

iBOWIMG \cite{zhou2015simple} &&& 55.7 & 76.5 & 35.0 & 42.6 & - &55.9& 78.7& 36.0& 43.4 & 62.0\\
DPPnet \cite{noh2016image} &&& 57.2 & 80.7 & 37.2 & 41.7 & - &57.4& 80.3& 36.9& 42.2 & - \\
VQA team \cite{antol2015vqa} &&& 57.8 & 80.5 & 36.8 & 43.1 & 62.7 &58.2& 80.6& 36.5& 43.7& 63.1\\
AYN \cite{malinowski2015ask}    &&& 58.4 & 78.4 & 36.4 & 46.3 & - & 58.4& 78.2& 36.3& 46.3& -\\
AMA \cite{wu2016ask}    &&& 59.2 & 81.0 & 38.4 & 45.2 & - &59.4& 81.1& 37.1& 45.8& -\\
DMN+ \cite{xiong2016dynamic} &&& 60.3 & 80.5 & 36.8 & 60.3 & - &60.4&-&-&-&- \\
MCB \cite{fukui2016multimodal}   && & 61.1 & 81.7 & 36.9 & 49.0 & - &61.1& 81.7& 36.9& 49.0& -\\
MRN \cite{kim2016multimodal}   && & 61.7 & \textbf{82.3} & \textbf{38.9} & 49.3 & - & 61.8& 82.4& 38.2& 49.4& 66.3 \\
MFB (Ours) &&& \textbf{62.2}& 81.8 & 36.7 & \textbf{51.2}& \textbf{67.2} &-&-&-&-&-\\
\hline
SMem \cite{xu2016ask}  &$\checkmark$& & 58.0 & 80.9 & 37.3 & 43.1 & - &58.2& 80.9& 37.3& 43.1&-\\
NMN \cite{andreas2016neural}   & $\checkmark$& & 58.6 & 81.2 & 38.0 & 44.0 & - &58.7& 81.2& 37.7& 44.0&-\\
SAN \cite{yang2016stacked}  & $\checkmark$&  & 58.7 & 79.3 & 36.6 & 46.1 & - & 58.9 &-&-&-&-\\
FDA \cite{ilievski2016focused}   & $\checkmark$& & 59.2 & 81.1 & 36.2 & 45.8 & -& 59.5&-&-&-&-\\
DNMN \cite{andreas2016learning} & $\checkmark$&  & 59.4 & 81.1 & 38.6 & 45.4 & - &59.4&-&-&-&-\\
HieCoAtt \cite{lu2016hierarchical}& $\checkmark$&& 61.8& 79.7 & 38.7 & 51.7 & 65.8 &62.1&-&-&-&-\\
RAU \cite{noh2016training} & $\checkmark$&& 63.3 & 81.9 & 39.0 & 53.0 & 67.7 &63.2& 81.7& 38.2& 52.8& 67.3\\
MCB+Att \cite{fukui2016multimodal} & $\checkmark$&& 64.2& 82.2 & 37.7 & 54.8 & - &-&-&-&-&-\\
DAN \cite{nam2016dual} & $\checkmark$&& 64.3& 83.0 & {39.1} & 53.9 & 69.1 & 64.2& 82.8& 38.1& 54.0&69.0\\
MFB+Att (Ours) & $\checkmark$& & 64.6 & 82.5& {38.3}& 55.2& 69.6 &-&-&-&-&-\\
MFB+CoAtt (Ours)  & $\checkmark$& & \textbf{65.1}& {83.2} & 38.8& \textbf{55.5}& \textbf{70.0} & -&-&-&-&-\\
\hline
MCB+Att+GloVe \cite{fukui2016multimodal} & $\checkmark$& $\checkmark$& 64.7 & 82.5 & 37.6 & {55.6}& - &-&-&-&-&-\\
MLB+Att+StV \cite{kim2016hadamard} & $\checkmark$& $\checkmark$ & 65.1 & \textbf{84.1} & 38.2 & 54.9 &- & 65.1& 84.0& 37.9& 54.8& 68.9 \\
MFB+CoAtt+GloVe (Ours) & $\checkmark$& $\checkmark$ & \textbf{65.9} & {84.0} & \textbf{39.8} & \textbf{56.2}& \textbf{70.6} &{65.8}& {83.8}& \textbf{38.9}& {56.3}& {70.5}\\
\hline
MCB+Att+GloVe+VG \cite{fukui2016multimodal} &$\checkmark$&$\checkmark$& 65.4& 82.3& 37.2& 57.4&-&-&-&-&-&-\\
MLB+Att+StV+VG \cite{kim2016hadamard} &$\checkmark$&$\checkmark$& 65.8& 83.9& 37.9& 56.8&-&-&-&-&-&-\\
MFB+CoAtt+GloVe+VG (Ours) & $\checkmark$& $\checkmark$ & \textbf{66.9}& \textbf{84.1}& \textbf{39.1}& \textbf{58.4} &\textbf{71.3}&\textbf{66.6}&\textbf{84.2}&38.1&\textbf{57.8}&\textbf{71.4}\\
\end{tabular}
\vspace{-10pt}
\end{table*}

First, MFB significantly outperforms MCB and MLB. With 5/6 parameters, MFB($k=5,o=1000$) achieves about a 1$\%$ accuracy improvement compared with MCB. Moreover, with only 1/3 parameters , MFB($k=5,o=200$) obtains similar results to MCB. These characteristics allows us to train our model on a memory limited GPU with larger batch-size. Furthermore,  the validation accuracy of MCB suffers from overfitting with the high-dimensional output features. In comparison, the performance of our MFB model is relatively robust.

Second, when $ko$ is fixed to a constant, e.g., 5000, the number of factors $k$ affects the performance. Increasing $k$ from 1 to 5, produces a 0.5$\%$ performance gain. When $k=10$, the performance has approached saturation. This phenomenon can be explained by the fact that a large $k$ corresponds to using a large window to sum pool the features, which can be treated as a compressed representation and may loss some information. When $k$ is fixed, increasing $o$ does not produce further improvements. This suggests that high-dimensional output features may be easier to overfit. Similar results can be seen in \cite{fukui2016multimodal}. In summary, $k=5$ and $o=1000$ may be a suitable combination for our MFB model on the VQA dataset, so we use these settings in our follow-up experiments.

Finally, both the power and $\ell_2$ normalization benefit MFB performance. Power normalization results in about 0.5$\%$ improvement and $\ell_2$ normalization, perhaps surprisingly, results in about $3\%$ improvement. Results without $\ell_2$ and power normalizations were also reported in \cite{antol2015vqa} and are similar to those reported here. To explain why normalization are so important, we randomly choose one typical neuron from the MFB output feature before normalization to illustrate how its distribution evolves over time in Figure \ref{fig:percentile}. It can be seen that the standard MFB model (with both normalizations) leads to the most stable neuron distribution and without the power normalization, about 10,000 iterations are needed to achieve stabilization. Without the $\ell_2$ normalization, the distribution varies seriously over the entire training course. This observation is consistent with the results shown in Table \ref{table:base}.

\subsection{Comparison with State-of-the-art}

Table \ref{table:sota} compares our approaches with the current state-of-the-art. The table is split into four parts over the rows: the first summarizes the methods without introducing the attention mechanism; the second includes the methods with attention; the third illustrates the results of approaches with external pre-trained word embedding models, e.g., GloVe \cite{pennington2014glove} or Skip-thought Vectors (StV) \cite{kiros2015skip}; and the last includes the models trained with the external large-scale Visual Genome dataset \cite{krishna2016visual} additionally. To best utilize model capacity, the training data set is augmented so that both the train and val splits are used as the training set, result in about $1\%\sim2\%$ overall accuracy improvement on the OE task. Also, to better understand the question semantics, pre-trained GloVe word vectors are concatenated with the learned word embedding. The MFB model corresponds to the MFB baseline model. The MFB+Att model indicates the model that replaces the MCB with our MFB in the MCB+Att model \cite{fukui2016multimodal}. The MFB+CoAtt model represents the network shown in Figure \ref{fig:mfb_coatt}.


From Table \ref{table:sota}, we have the following observations:

First, the model with MFB outperforms other comparative approaches significantly. The MFB baseline outperforms all other existing approaches without the attention mechanism for both the OE and MC tasks, and even surpasses some approaches with attention. When attention is introduced, MFB+Att consistently outperforms current next-best model MCB+Att, highlighting the efficacy and robustness of the proposed MFB.
\begin{figure*}
\begin{center}
\includegraphics[width=0.98\textwidth]{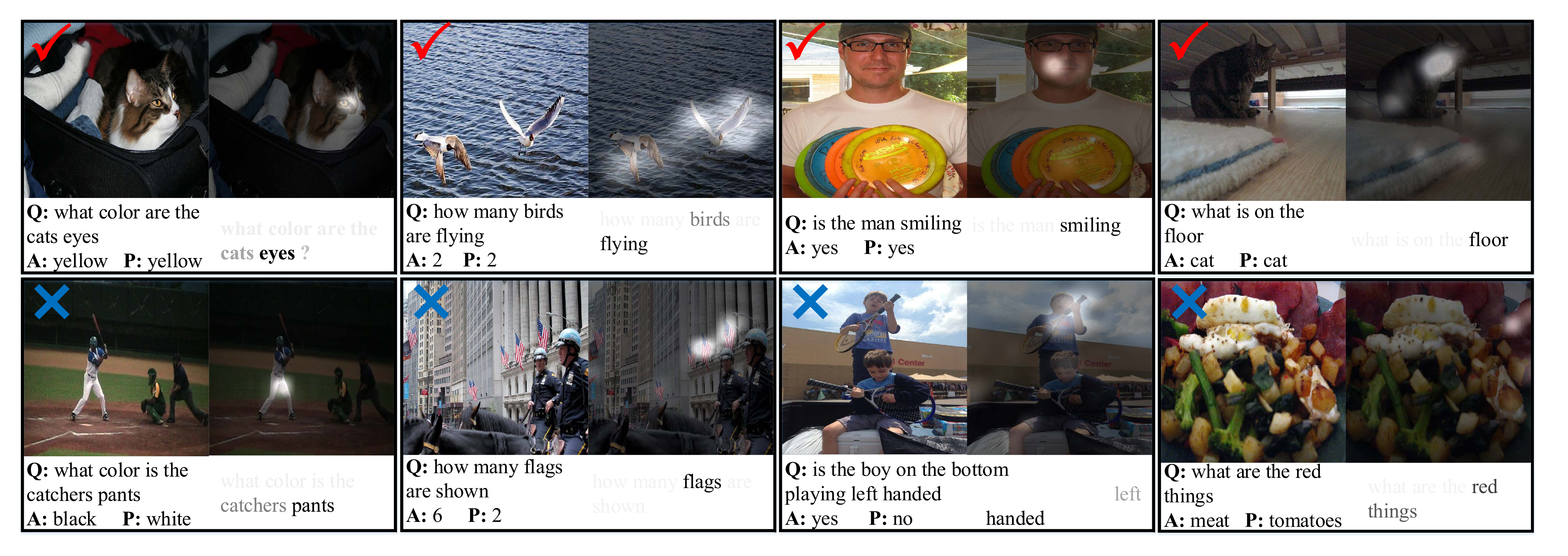}
\caption{Typical examples of the learned image and question of the MFB+CoAtt+GloVe model. The top row shows four examples of four correct predictions while the bottom row shows four incorrect predictions. For each example, the query image, question (Q), answer (A) and prediction (P) are presented from top to bottom; the learned image and question attentions are presented next to them. The brightness of images and darkness of words represent their attention weights.}
\label{fig:mfb_visualizayion}
\end{center}
\vspace{-10pt}
\end{figure*}

Second, the co-attention model further improve the performance over the attention model with only considering the image attention. By introducing co-attention learning, MFB+CoAtt delivers a 0.5$\%$ improvement on the OE task compared with the MFB+Att model in terms of overall accuracy, indicating the additional benefits of the co-attention learning framework.

Finally, with the external pre-trained GloVe model and the Visual Genome dataset, the performance of our models are further improved. The MFB+CoAtt+GloVe+VG model significantly outperforms the best reported results with a single model on both the OE and MC task.

In Table \ref{table:test-std}, we compare our approach with the state-of-the-art methods with model ensemble. Similar with \cite{fukui2016multimodal,kim2016hadamard}, we train 7 individual MFB+CoAtt+GloVe models and average the prediction scores of them. Four of the seven models additionally introduce the Visual Genome dataset \cite{krishna2016visual} into the training set.  For fair comparison, only the published results are demonstrated. From Table \ref{table:test-std}, the ensemble of MFB models outperforms the next best approach by 1.5$\%$ on the OE task and by 2.2$\%$ on the MC task respectively. Finally, compared with the results obtained by human, there is still a lot of room for the improvement to approach the human-level.

\begin{table}
\centering
\caption{Comparison with the state-of-the-art results (with model ensemble) on the test-standard set of the VQA dataset. The best results are bolded.}
\small
\label{table:test-std}
\begin{tabular}{lccccc}
{Model} & \multicolumn{4}{c}{{OE}} & {MC}\\
\hline
 & All & Y/N & Num & Other & All \\
\cline{2-5}
HieCoAtt \cite{lu2016hierarchical} & 62.1& 80.0& 38.2& 52.0 & 66.1\\
RAU \cite{noh2016training} & 64.1 & 83.3 & 38.0 & 53.4 & 68.1\\
7 MCB models \cite{fukui2016multimodal} & 66.5 & 83.2& 39.5& 58.0& 70.1 \\
7 MLB models \cite{kim2016hadamard} & 66.9 & 84.6& 39.1 &57.8 & 70.3\\
7 MFB models (Ours) & \textbf{68.4} & \textbf{85.6} & \textbf{40.6} & \textbf{59.8} & \textbf{72.5} \\
\hline
Human \cite{antol2015vqa} & 83.3& 95.8& 83.4& 72.7& 91.5 \\
\end{tabular}
\vspace{-10pt}
\end{table}

To better demonstrate the effects of co-attention learning, in Figure \ref{fig:mfb_visualizayion} we visualize the learned question and image attentions of some examples from the validation set. The examples are randomly picked from different question types. It can be seen that the learned question and image attentions are usually closely focus on the key words and the most relevant image regions. From the incorrect examples, we can also draw conclusions about the weakness of our approach, which are perhaps common to all VQA approaches: 1) some key words in the question are neglected by the question attention module, which seriously affects the learned image attention and final predictions (e.g., the word \emph{catcher} in the first example and the word \emph{bottom} in the third example); 2) even the intention of the question is well understood, some visual contents are still unrecognized (e.g., the \emph{flags} in the second example) or misclassified (the \emph{meat} in the fourth example), leading to the wrong answer for the counting problem. These observations are useful to guide further improvements for VQA in the future.

\section{Conclusions}\label{sec:conclusion}
In this paper, we develop a Multi-modal Factorized Bilinear pooling (MFB) approach to fuse multi-modal features for the VQA task. Compared with existing bilinear pooling methods, the MFB approach achieves significant performance improvement for the VQA task. Based on MFB, we design a network architecture with co-attention learning that achieves new state-of-the-art performance on the real-world VQA dataset. This explorations of multi-modal bilinear pooling and co-attention learning are applicable to a wide range of tasks involving multi-modal data.

\section{Acknowledgement}
This work was supported in part by the National Natural Science Foundation of China under Grants 61622205, 61602136 and 61472110, the Zhejiang Provincial Natural Science Foundation of China under Grant LR15F020002,  the Australian Research Council under Project FL-170100117,  DP-140102164, and LP-150100671.
\newpage
{\small
\bibliographystyle{ieee}
\bibliography{0675.bbl}
}

\end{document}